\documentclass[conference]{IEEEtran}
\IEEEoverridecommandlockouts

\usepackage{cite}
\usepackage{amsmath,amssymb,amsfonts}
\usepackage{graphicx}
\usepackage{textcomp}
\usepackage{xcolor}

\usepackage{amsthm}
\usepackage{amsfonts}
\usepackage{float}
\usepackage{xspace}
\usepackage{bm}
\usepackage{color}
\usepackage{tabu}
\usepackage{CJK}
\usepackage{booktabs}
\usepackage{xcolor}
\usepackage{multirow}
\usepackage{array}
\usepackage{graphicx}
\usepackage{url}
\usepackage{makecell}
\usepackage{subcaption}
\usepackage{algorithm}
\usepackage{algpseudocode}
\usepackage{pgfplots}
\usepackage{subcaption}

\def\BibTeX{{\rm B\kern-.05em{\sc i\kern-.025em b}\kern-.08em
    T\kern-.1667em\lower.7ex\hbox{E}\kern-.125emX}}

\begin{document}

\title{Do LLMs Need Architectural Changes for Simultaneous Speech Translation? A Prefix-to-Prefix Data Driven Approach\\
}

\author{\IEEEauthorblockN{Junkun Chen, Jian Xue, Ming Tang, Abdel Heba, Hoda Gholami, Ruchao Fan, Jinyu Li\\
\IEEEauthorblockN{\{junkunchen,jinyli\}@microsoft.com}}
\IEEEauthorblockA{Microsoft, USA}}

\maketitle

\maketitle

\begin{abstract}
Simultaneous speech translation (SimulST) requires incremental translation under strict latency constraints, yet remains challenging for decoder-only LLM systems due to limited context and cross-lingual reordering.
Recent approaches often introduce architectural changes or explicit read/write policies to control output timing, which can be brittle in conversational speech where segmentation boundaries are ambiguous.
We present a simple data-driven alternative: fixed-length chunks for cumulative streaming decoding with a rewind-based committed prefix, and teacher-labeled prefix-to-prefix (P2P) targets with bounded waiting for fine-tuning, yielding CSSEL-P2P, where CSSEL is our proposed chunked streaming speech encoder LLM.
In our in-house conversational speech evaluation, CSSEL-P2P improves streaming quality by +1.54 COMETKiwi over the CSSEL streaming baseline at comparable latency (+0.15s Average Lagging), suggesting effective SimulST without architectural changes via P2P supervision.

\end{abstract}


\section{Introduction}
\label{sec:intro}

Simultaneous speech translation (SimulST) enables real-time cross-lingual communication in applications such as meetings, live events, and customer support~\cite{ma2020simulmt,ma2021streaming}.
In recent years, large language models (LLMs) have advanced rapidly and demonstrated strong performance for \emph{offline} speech translation when full utterance context is available~\cite{achiam2023gpt,touvron2023llama,liu2024deepseek,team2025gemma,yang2025qwen3}.
Despite this progress, building LLM-based SimulST systems remains challenging because translation must be produced incrementally under strict latency constraints.
The core objective is to achieve a favorable quality--latency trade-off.

A fundamental difficulty in SimulST is cross-lingual \emph{reordering}.
Early speech prefixes often do not contain enough information to determine a faithful and fluent target prefix, especially for reordering-heavy directions.
Consequently, effective SimulST requires not only efficient decoding, but also learning \emph{when} and \emph{what} to translate under limited context.
This observation has motivated a large body of work on explicit read/write control and latency-aware modeling.
For example, fixed-latency strategies such as wait-$k$~\cite{ma2019stacl} provide a simple policy baseline, while learned policies and monotonic attention variants enable adaptive schedules~\cite{arivazhagan2019monotonic,chiu2017monotonic,ma2019monotonic}.

In speech translation, additional complexity arises from the lack of clear segmentation boundaries in conversational audio.
Disfluencies, pauses, interruptions, and variable speaking rates 
make fine-grained boundary decisions usually ambiguous and error-prone.
As a result, approaches that rely on explicit segmentation, continuous trigger signals, or learned read/write policies can be brittle to deploy in realistic dialog settings.
A related line of work adopts \emph{re-translation}, repeatedly translating a growing input from scratch and allowing revisions of intermediate outputs~\cite{arivazhagan2020re,arivazhagan2020respeech}.
More recently, LLM-based SimulST systems have reformulated SimulST as an \emph{interleaved generation} problem with explicit read/write tokens and introduced dedicated policy heads to predict output timing.
\cite{fu2025efficient,ouyang2025infinisst,ouyang2024fasst}.

This landscape raises a simple question that directly motivates our work.
\textbf{Do decoder-only LLMs really need architectural changes to work well for SimulST?}
We argue that, under a robust streaming interface, the dominant bottleneck is often \emph{data} rather than \emph{architecture}.
Specifically, an offline-trained LLM is not trained to produce \emph{commit-ready} partial translations from prefix-only speech inputs.
Instead of adding policy modules or designing complex boundary detectors, we adopt a simple and direct strategy.
We process the incoming speech using fixed-length chunks of $\Delta$ seconds and perform cumulative streaming decoding.
This fixed-chunk interface avoids the need to predict uncertain speech boundaries and provides a stable deployment abstraction.
We further introduce a rewind boundary of length $k$ to determine a committed prefix at each step, and only the committed prefix is emitted to the user.

Crucially, we teach the model what to output at each prefix by constructing prefix-to-prefix (P2P) supervision with a teacher LLM.
We obtain chunk-level transcriptions via forced alignment \emph{for labeling only}.
We then prompt a teacher LLM to plan a bounded-waiting schedule over the full chunk sequence, producing incremental translation prefixes that are smoothly extendable under reordering.
These teacher-labeled prefixes define training targets for speech-prefix-to-prefix fine-tuning of a decoder-only LLM-based speech translation model.
We evaluate on in-house conversational speech and report translation quality with COMETKiwi~\cite{rei2022cometkiwi} alongside latency with AL
\cite{ma2020simuleval}.
The results show that P2P fine-tuning substantially improves streaming quality at comparable latency, with larger gains on reordering-heavy directions.

Our contributions are as follows.
\begin{itemize}
\item We propose a simple fixed-chunk streaming recipe for decoder-only LLM-based speech translation, with explicit knobs (chunk size $\Delta$ and rewind length $k$) to navigate the quality--latency frontier.
\item We introduce a teacher-driven prefix-to-prefix data construction method that produces incremental, commit-ready targets under bounded waiting. Fine-tuning on these targets teaches the model what to commit from prefix-only inputs and strengthens the streaming decoding behavior.
\item We demonstrate consistent improvements on in-house conversational speech across multiple directions in quality, and provide trade-off analyses over $\Delta$ and $k$.
\end{itemize}

\section{Preliminaries}
\label{sec:prelim}

\subsection{Offline vs.\ Streaming Speech Translation}
\label{sec:prelim:offline_streaming}
Let $s$ denote an utterance-level speech waveform, 
$x$ its source-language transcription, 
and $y$ the target-language translation. 
In \textbf{offline} ST, 
the full utterance $s$ is available before decoding, and the system outputs a translation $y$ using complete acoustic and linguistic context. 
This setting allows the model to defer decisions and resolve ambiguity using future evidence (e.g., later words or clauses).

In SimulST, the input arrives incrementally. 
At time $t$, only a speech prefix $s_{\le t}$ is observable, 
and the system produces an evolving translation prefix $y_{\le t}$ without access to any future speech. 
The model must therefore trade off \emph{latency} and \emph{quality}: emitting early improves simultaneity but may be error-prone under limited context; delaying output improves accuracy but increases latency.


\subsection{Reordering Across Audio, Transcription, and Translation}
\label{sec:prelim:reordering}
A central challenge in SimulST is \textbf{reordering} between the source and target languages~\cite{ren2020simulspeech,omachi2023align}. While speech $s$ unfolds temporally and its transcription $x$ is revealed roughly left-to-right, the target translation $y$ may require a different word order to be grammatical and natural. This implies that a target prefix is not necessarily a monotonic function of the source prefix.

Formally, consider a sequence of growing source transcription prefixes $\{x_{\le 1}, x_{\le 2}, \ldots\}$ derived from $\{s_{\le t_1}, s_{\le t_2}, \ldots\}$. Under strong reordering, early source prefixes may not contain enough information to determine a faithful and fluent translation prefix. Premature commitment can lead to:
\begin{itemize}
  \item \textbf{semantic errors} due to unresolved ambiguity (e.g., polysemy or missing arguments),
  \item \textbf{grammatical infelicities} from incomplete structures,
\end{itemize}
Conversely, always waiting for sufficient context results in high latency and poor simultaneity. Therefore, an effective SimulST system must learn when it is safe to \emph{commit} translation content and when it should \emph{wait} for additional context. 

\subsection{Motivation for Chunk-Based Streaming and Prefix-to-Prefix Supervision}
\label{sec:prelim:motivation}
To achieve streaming decoding for ST, a common practice is to process the speech stream in fixed increments. At step $i$, the system observes a speech prefix $s_{\le t_i}$ (or equivalently, the accumulation of the first $i$ chunks) and outputs an updated translation~\cite{chen2021direct}. This motivates a \textbf{chunk-based streaming decoding} approach, where the model is repeatedly invoked as the available speech context grows.

However, chunk-based decoding alone does not solve the reordering problem: without appropriate training signals, the model may either revise earlier translations aggressively or delay excessively. A natural idea is to train the model with \textbf{prefix-to-prefix} examples, where each training instance maps a source prefix to a target prefix that is valid \emph{without using future source context}. Such supervision teaches the model to produce stable, monotonic outputs under streaming constraints.

Crucially, constructing P2P targets is non-trivial. Naively truncating an offline full-sentence translation to obtain target prefixes can produce unnatural partial outputs that implicitly rely on future context (especially under reordering)~\cite{chen2021improving}, leading to inconsistent training objectives. This motivates a more principled target construction strategy that explicitly models bounded waiting and produces smoother incremental translations.

In the next section, we introduce (i) a simple cumulative chunk streaming decoding method with prefix anchoring to improve stability and quality, and (ii) a teacher-driven P2P data construction and fine-tuning procedure that strengths LLM-based simultaneous speech translation under reordering.

\section{Method}
\label{sec:method}

\begin{figure*}[t]

    \centering
    \includegraphics[width=1\linewidth]{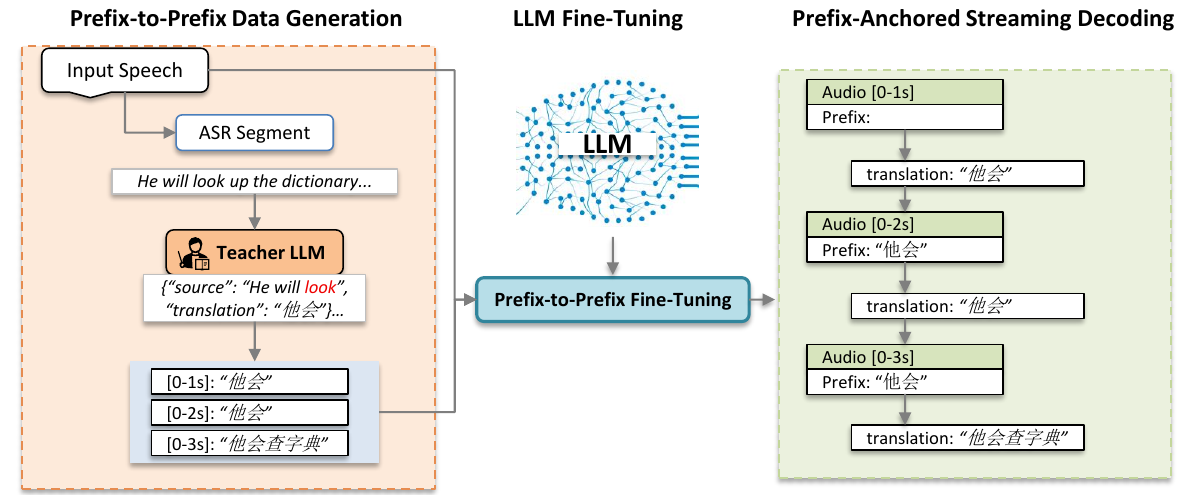}
    \caption{Prefix-to-prefix supervision enables decoder-only LLMs for SimulST without architectural changes.
    A teacher LLM plans incremental, commit-ready translations from chunk-level transcriptions. The student LLM model is fine-tuned on these prefix targets and decoded online by anchoring on the emitted prefix while consuming cumulative audio chunks.}
    \label{fig:pipeline}
    \end{figure*}
    
We propose processing streaming speech in fixed increments with chunk size $\Delta$ seconds. At step $i$, the observed speech prefix is $s_{\le t_i}$ where $t_i=i\Delta$, and the system outputs a target translation prefix $\hat{y}_{\le i}$. We target a \textbf{no-revision} streaming interface: once a translation prefix is emitted at step $i$, it must remain unchanged for all future steps.

\subsection{Hard Prefix-Anchored Chunk Streaming Decoding}
\label{sec:method:decoding}
Our inference procedure adopts \textbf{cumulative chunk decoding}. At each step $i$, we invoke the model on the entire speech prefix $s_{\le t_i}$. Let $y^{\text{emit}}_{i}$ as prefix committed at the previous step. We perform \textbf{hard prefix-anchored decoding} by forcing the decoder to start from $y^{\text{emit}}_{i}$ and only generate an extension:
\begin{equation}
\hat{y}_{\le i} = \mathrm{Decode}(s_{\le t_i};\, \mathrm{prefix}=y^{\text{emit}}_{i}).
\label{eq:hard_prefix_decode}
\end{equation}

This ensures monotonic growth of the output and eliminates revisions by construction.

\subsubsection{Rewind-based prefix anchoring.}
To avoid locking in unstable boundary decisions, at step $i$ we rewind the last $k$ tokens of the predicted output and only treat the remaining prefix as committed:
\begin{equation}
\hat{y}'_{\le i-1} = \mathrm{rewind}(\hat{y}_{\le i-1}, k),
\end{equation}
\begin{equation}
\hat{y}_{\le i} = \mathrm{Decode}\!(s_{\le t_i};~\mathrm{prefix}=\hat{y}'_{\le i-1}).
\end{equation}
The rewind operation withholds the last $k$ tokens from commitment rather than revising user-visible output. These tokens may be generated internally, but they are not displayed to the user and are not used as the hard prefix in the next step. Only the remaining prefix is emitted and committed. At the next step, decoding is anchored on this committed prefix, so all user-visible tokens remain fixed, while the hidden suffix can be regenerated after additional speech context becomes available. The detailed method is described in Alg.~\ref{alg:streaming_decode_commit_anchor}. 

After each step, the emitted prefix is a monotonic user-visible prefix and satisfies
$y_{i-1}^{\mathrm{emit}} \preceq y_i^{\mathrm{emit}}$.
The last $k$ generated tokens form a hidden editable buffer and are excluded from the committed prefix. Therefore, the system remains revision-free from the user's perspective, while still avoiding premature commitment near the hypothesis boundary.

\subsubsection{Efficiency.}
Adjacent steps share substantial overlap in the speech prefix. We can reuse cached computations when available, including encoder representations of historical speech segments and decoder KV states for the processed audio prefix, reducing recomputation overhead in streaming settings.

\subsection{Teacher LLM Labeling for Commit-Ready Prefix Targets}
\label{sec:method:teacher}
Rewind-based decoding improves robustness at chunk boundaries, but the model can still make suboptimal early commitments if it has never been trained under prefix-only constraints. To address this, we construct \textbf{commit-ready} prefix targets using a teacher LLM, and fine-tune the student model accordingly.

\subsubsection{Chunk-level transcriptions for labeling.}
Using forced alignment, we obtain chunk-level source transcriptions $\{x_1,\ldots,x_T\}$, each corresponding to a fixed speech interval. These chunk transcriptions are used only for teacher labeling: they provide a clean, text-based representation of what information is available at each time step.

\subsubsection{Teacher planning with bounded waiting.}
We prompt a strong instruction-following LLM as a \emph{teacher} to mimic a professional simultaneous translator. During annotation, we provide the teacher with the entire sequence of chunks $\{x_1,\ldots,x_T\}$, enabling global planning. The teacher is instructed to decide, for each chunk $i$, whether to translate immediately based only on the available prefix context (previous chunks and the current chunk), or to \emph{wait} (output an empty translation) when context is insufficient. To bound latency, we impose a maximum waiting budget of $K$ chunks: if the teacher has waited for $K$ consecutive chunks, it must produce a translation at the next chunk. 
We set $K=3$ in our experiment.

\noindent The teacher outputs the corresponding translation per chunk:
\begin{equation}
\left\{(\texttt{source}=x_i,~\texttt{translation}=\tilde{t}_i)\right\}_{i=1}^{T},
\label{eq:teacher_output}
\end{equation}
where $\tilde{t}_i$ is either a target translation or an empty string when waiting. The commit-ready prefix at step $i$ is defined as:
\begin{equation}
\tilde{y}_{\le i}=\mathrm{concat}(\tilde{t}_1,\ldots,\tilde{t}_i).
\label{eq:teacher_prefix}
\end{equation}
This produces incremental targets that are explicitly designed to be \emph{smoothly extendable} under cross-lingual reordering.

\subsubsection{Teacher Labeling Constraints}

Although the teacher observes the full chunk sequence during
annotation, it is instructed to produce targets that simulate a
causal simultaneous translator. For each step, the teacher must
decide whether the available source prefix is sufficient for a
faithful target continuation. If the current prefix is ambiguous
or would force an unstable target-side word order, the teacher
is allowed to wait by emitting an empty string. The maximum
waiting budget $K$ prevents the teacher from deferring
translation indefinitely. 
This design uses the teacher's global knowledge to construct
a coherent schedule, while still producing prefix targets that
respect a bounded-latency streaming interface. The resulting
labels encode both lexical translation and emission timing,
but they require no architectural changes to the student model.

\begin{algorithm}[t]
\caption{Rewind-based Streaming Decoding}
\label{alg:streaming_decode_commit_anchor}
\begin{algorithmic}[1]
\Require Speech stream $s$; chunk size $\Delta$; steps $T$; rewind length $k$
\Ensure Emitted outputs $\{y^{\text{emit}}_1,\ldots,y^{\text{emit}}_T\}$

\State $\hat{y}_{0} \gets \emptyset$
\State $y^{\text{emit}}_{0} \gets \emptyset$
\State $\mathcal{C} \gets \emptyset$ \Comment{optional cache state}

\For{$i \gets 1$ to $T$}
  \State $t_i \gets i\Delta$
  \State Observe cumulative speech prefix $s_{\le t_i}$

  \State $y^{\text{anchor}}_{i-1} \gets y^{\text{emit}}_{i-1}$  \Comment{emitted text as the anchored prefix}

  \State $(\hat{y}_{i}, \mathcal{C}) \gets \mathrm{Decode}(s_{\le t_i};\, \mathrm{prefix}=y^{\text{anchor}}_{i-1},\, \mathcal{C})$

  \If{$i < T$}
    \State $y^{\text{emit}}_{i} \gets \mathrm{rewind}(\hat{y}_{i}, \min(k, |\hat{y}_{i}|))$
  \Else
    \State $y^{\text{emit}}_{i} \gets \hat{y}_{i}$ \Comment{emit full translation at the end}
  \EndIf

  \State Emit $y^{\text{emit}}_{i}$
\EndFor

\State \Return $\{y^{\text{emit}}_1,\ldots,y^{\text{emit}}_T\}$
\end{algorithmic}
\end{algorithm}

\subsection{Speech-to-Text Prefix-to-Prefix Fine-Tuning}
\label{sec:method:finetune}
We finetune the LLM-based speech translation model with the teacher-constructed commit-ready prefix-to-prefix data. For each utterance and each step $i$, we create a training instance mapping the speech prefix to the teacher-defined translation prefix, $\{s_{\le t_i} ,\tilde{y}_{\le i}\}$.

We optimize standard token-level cross entropy to maximize the likelihood of $\tilde{y}_{\le i}$ conditioned on $s_{\le t_i}$. Notably, the chunk transcriptions $\{x_i\}$ are not used as inputs to the student; the student learns to generate commit-ready prefixes directly from partial speech.

At inference time, we apply the rewind-based decoding in Section~\ref{sec:method:decoding}. Both training and inference operate on growing prefixes and aim to produce stable, monotonic outputs, making the fine-tuning objective well aligned with the streaming deployment setting. The whole pipeline is shown in Fig.~\ref{fig:pipeline}.

\subsection{Training--Inference Alignment}

A key design principle of our approach is to align the
fine-tuning target with the deployment-time streaming
interface. In standard offline speech translation training, the
model is optimized to generate a complete target sentence
given the full source utterance. When the same model is used
on partial speech prefixes, the target prefix implied by the
offline reference may not be commit-ready: it may require
future source context, contain premature lexical choices, or
follow a word order that is only valid after observing the full
utterance.

Our P2P objective instead trains the model on pairs
$(s_{\leq t_i}, \tilde{y}_{\leq i})$, where the target prefix
$\tilde{y}_{\leq i}$ is explicitly constructed to be safe to emit
under bounded waiting. This matches the inference-time
constraint that the emitted prefix should grow monotonically
and should not require substantive revision. As a result, P2P
fine-tuning does not merely adapt the model to shorter audio
inputs; it teaches the model a streaming behavior: when to
emit, when to wait, and how to produce prefixes that remain
extendable under future context.

\section{Experiments}
\label{sec:exp}

\subsection{Model and Training Data}
\label{sec:exp:model_data}

\subsubsection{Backbone and architecture.}
Our LLM backbone is Phi-4-MM~\cite{abouelenin2025phi}. Different from the original design, we processed the audio input with a \textbf{chunked streaming speech encoder}~\cite{xue2022large} to better support incremental speech processing. 
The audio encoder processes speech with an 80 ms frame rate after temporal subsampling. We refer to the resulting model as \textbf{CSSEL} (\underline{C}hunked \underline{S}treaming \underline{S}peech \underline{E}ncoding \underline{L}LM).

\textbf{Streaming speech encoder} The encoder takes 80-dimensional log-Mel filterbank features extracted with a 10 ms frame shift. A convolutional front-end performs temporal subsampling, producing one acoustic frame every 80 ms, which are then processed by a stack of 12 Conformer blocks~\cite{gulati20_interspeech}. Each block uses an attention dimension of 1280 with 20 attention heads, a 4480-dimensional feed-forward layer, a causal depthwise convolution with kernel size 3, and Swish activations; relative positional encoding is used in the self-attention. To enable streaming, all convolutions are made causal and the self-attention is restricted by a \emph{chunk-wise attention mask}: frames are grouped into fixed-size chunks (10 frames, i.e.\ 800 ms), and each frame attends only to frames within its own chunk and a bounded left context of 7 preceding chunks, with no access to future frames beyond the current chunk. This chunked masking bounds the look-ahead latency while still allowing a limited right context within a chunk, giving a controllable trade-off between recognition accuracy and streaming latency.

\textbf{Offline Training data}:
CSSEL inherits the same language coverage as its Phi-4-MM. We first train CSSEL in the standard offline setting using a large-scale mixture of speech translation directions. Specifically, our offline corpus includes:
\begin{itemize}
    \item \textbf{20k hours} of $\mathrm{X}\rightarrow\mathrm{En}$ speech translation pairs,
    \item \textbf{20k hours} of $\mathrm{En}\rightarrow\mathrm{Y}$ speech translation pairs,
    \item \textbf{3k hours} of direct $\mathrm{X}\rightarrow\mathrm{Y}$ speech translation pairs.
\end{itemize}

\textbf{Prefix-to-prefix fine-tuning data}:
To train CSSEL for simultaneous translation, we additionally collect prefix-to-prefix supervision on a smaller, carefully constructed dataset. We sample 200 hours for each language pair of $\mathrm{X}\rightarrow\mathrm{En}$ and $\mathrm{En}\rightarrow\mathrm{X}$. We segment each utterance using a fixed chunk size of 4 seconds and obtain chunk-level transcriptions via forced alignment~\cite{rousso2024tradition}. We then use \textbf{GPT-o3-mini}\footnote{https://openai.com/index/openai-o3-mini/} as a teacher to annotate prefix-to-prefix translation targets following the bounded-waiting protocol described in Section~\ref{sec:method:teacher}. These teacher-labeled prefix pairs are used for the prefix-to-prefix fine-tuning stage.
We adopt a two-stage training recipe: (i) offline training on the large-scale corpus to obtain \textbf{CSSEL}, 
followed by (ii) P2P fine-tuning on the teacher-labeled 4s-chunk dataset to induce stable streaming behavior, yielding \textbf{CSSEL-P2P}.



We evaluated this work on in-house conversational speech corpora collected from real dialog scenarios. 
We report translation quality using COMETKiwi\footnote{https://huggingface.co/Unbabel/wmt22-cometkiwi-da} for all directions and all models for a consistent comparison.

\begin{table}[t]
\resizebox{\linewidth}{!}{%
\centering
\small
\setlength{\tabcolsep}{4pt}
\begin{tabular}{lccccc}
\toprule
\multirow{2}{*}{\textbf{Direction}} &
\multicolumn{3}{c}{\textbf{Offline}} &
\multicolumn{2}{c}{\textbf{Streaming ($\Delta=4s$)}} \\
\cmidrule(lr){2-4}\cmidrule(lr){5-6}
& \textbf{Phi-4-MM} & \textbf{Gemini 3.0} & \textbf{CSSEL} & \textbf{CSSEL} & \textbf{CSSEL-P2P} \\
\midrule
EN$\rightarrow$DE           & 68.89 & 67.29 & 69.01 & 68.34 & 69.37 \\
EN$\rightarrow$ES           & 67.88 & 66.99 & 68.31 & 67.97 & 68.62 \\
EN$\rightarrow$FR           & 68.11 & 67.75 & 68.82 & 68.08 & 69.56 \\
EN$\rightarrow$IT           & 70.75 & 69.42 & 70.38 & 69.63 & 70.95 \\
EN$\rightarrow$JA           & 71.98 & 74.51 & 72.41 & 69.46 & 73.42 \\
EN$\rightarrow$PT           & 66.09 & 65.45 & 66.14 & 65.69 & 67.08 \\
EN$\rightarrow$ZH      & 66.96 & 68.36 & 67.44 & 66.24 & 68.98 \\
\midrule
DE$\rightarrow$EN        & 71.00 & 72.62 & 72.30 & 71.57 & 72.24 \\
ES$\rightarrow$EN        & 76.44 & 78.23 & 78.07 & 77.52 & 77.69 \\
FR$\rightarrow$EN        & 71.10 & 75.30 & 74.56 & 74.31 & 74.51 \\
IT$\rightarrow$EN        & 75.32 & 74.92 & 74.91 & 74.32 & 74.49 \\
JA$\rightarrow$EN        & 74.32 & 76.71 & 74.68 & 73.53 & 74.34 \\
PT$\rightarrow$EN        & 66.73 & 73.28 & 71.20 & 70.42 & 72.58 \\
ZH$\rightarrow$EN      & 70.07 & 73.66 & 70.88 & 70.02 & 71.08 \\
\midrule
ES$\rightarrow$DE        & 68.11 & 67.44 & 68.24 & 67.21 & 68.45 \\
FR$\rightarrow$PT        & 68.22 & 70.05 & 69.09 & 68.61 & 70.07 \\
IT$\rightarrow$ZH   & 60.06 & 65.22 & 59.85 & 58.48 & 61.88 \\
ZH$\rightarrow$JA        & 61.03 & 70.40 & 61.07 & 59.82 & 63.76 \\
\midrule
\textbf{Avg.}               & 69.06 & 70.98 & 69.85 & 68.96 & 70.50 \\
\midrule
\textbf{AL (s)}  & - & - & - & 2.19 & 2.34 \\
\bottomrule
\end{tabular}
}
\caption{COMETKiwi performance across language directions on in-house conversational speech. Offline denotes full-utterance decoding, while streaming results use cumulative chunk decoding with $\Delta=4$s and rewind length $k=5$. AL reports average lagging in seconds for streaming systems. }
\label{tab:main_results_multi}
\end{table}


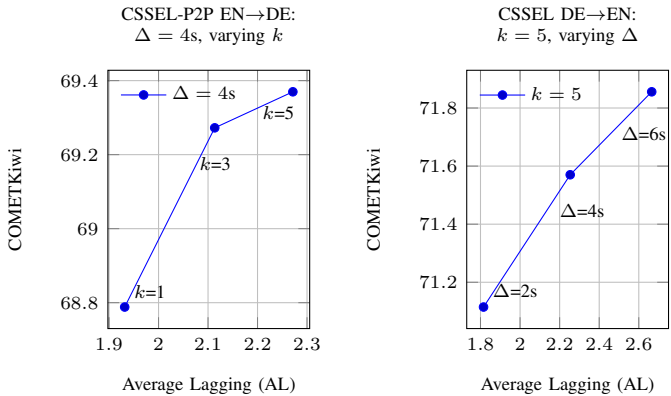
\begin{figure}[t]
\centering
\begin{tikzpicture}
\begin{axis}[
    width=0.48\linewidth,
    height=5.0cm,
    xlabel={Average Lagging (AL)},
    ylabel={COMETKiwi},
    title={CSSEL-P2P EN$\rightarrow$DE:\\ $\Delta=4$s, varying $k$},
    title style={align=center, font=\scriptsize},
    grid=both,
    tick label style={font=\scriptsize},
    label style={font=\scriptsize},
    title style={font=\scriptsize},
    legend style={font=\scriptsize, at={(0.02,0.98)},anchor=north west,draw=none,fill=none},
]
\addplot+[mark=*, mark size=1.6pt] coordinates {
    (1.9316, 68.78826)
    (2.1135, 69.27246)
    (2.2714, 69.36982)
};
\addlegendentry{$\Delta=4$s}

\node[anchor=south west, font=\scriptsize] at (axis cs:1.9316,68.78826) {$k$=1};
\node[anchor=south west, font=\scriptsize] at (axis cs:2.0635,69.13246) {$k$=3};
\node[anchor=south west, font=\scriptsize] at (axis cs:2.1914,69.26982) {$k$=5};
\end{axis}
\end{tikzpicture}
\hfill
\begin{tikzpicture}
\begin{axis}[
    width=0.48\linewidth,
    height=5.0cm,
    xlabel={Average Lagging (AL)},
    ylabel={COMETKiwi},
    title={CSSEL DE$\rightarrow$EN: \\ $k=5$, varying $\Delta$},
    title style={align=center, font=\scriptsize},
    grid=both,
    tick label style={font=\scriptsize},
    label style={font=\scriptsize},
    title style={font=\scriptsize},
    legend style={font=\scriptsize, at={(0.02,0.98)},anchor=north west,draw=none,fill=none},
]
\addplot+[mark=*, mark size=1.6pt] coordinates {
    (1.8154, 71.11468)
    (2.2534, 71.57014)
    (2.6653, 71.85581)
};
\addlegendentry{$k=5$}

\node[anchor=south west, font=\scriptsize] at (axis cs:1.8154,71.11468) {$\Delta$=2s};
\node[anchor=south west, font=\scriptsize] at (axis cs:2.1534,71.39014) {$\Delta$=4s};
\node[anchor=south west, font=\scriptsize] at (axis cs:2.4653,71.65581) {$\Delta$=6s};
\end{axis}
\end{tikzpicture}

\caption{Quality--latency trade-offs (COMETKiwi vs.\ AL) on EN$\rightarrow$DE and DE$\rightarrow$EN. Left: varying rewind length ($k$) at fixed $\Delta=4$s. Right: varying chunk size $\Delta$ at fixed $k$=5.}
\label{fig:de2en_tradeoff_lr}
\end{figure}

\subsubsection{Main Results}
\label{sec:exp:main}
In Tab~\ref{tab:main_results_multi}, we report COMETKiwi on our in-house conversational speech across multiple directions.
We include Phi-4-MM and Gemini 3.0 as \emph{reference} systems.
Since CSSEL is trained with an in-house data mixture and recipe that may differ in domain from the backbones, we focus our main comparisons on \textbf{CSSEL vs.\ CSSEL-P2P} under the same streaming configuration.
Under offline full-decoding, CSSEL achieves $69.85$ COMETKiwi on average, serving as a strong baseline.
Switching to streaming decoding with $\Delta=4$s reduces CSSEL to 68.96, which is expected under limited context and a no-revision (prefix-anchored) interface.
P2P finetuning substantially improves streaming decoding, CSSEL-P2P reaches 70.50, a \textbf{+1.54} gain over CSSEL streaming, while average lagging increases only from 2.19s to 2.34s (\textbf{+0.15}s).
Gains are consistent across directions and are larger on reordering-heavy pairs, e.g., EN$\rightarrow$JA (+3.96), EN$\rightarrow$ZH (+2.74), and ZH$\rightarrow$JA (+3.94).
Notably, our prefix-to-prefix supervision is constructed only for X$\rightarrow$EN and EN$\rightarrow$X directions; however, we still observe improvements on X$\rightarrow$Y directions that are not directly supervised (e.g., ZH$\rightarrow$JA). This suggests that P2P finetuning primarily shapes general streaming decoding behavior, rather than language-pair-specific tuning.

\subsubsection{Quality--Latency Trade-offs}
\label{sec:exp:tradeoff}
Figure~\ref{fig:de2en_tradeoff_lr} analyzes quality--latency trade-offs on EN$\rightarrow$DE and DE$\rightarrow$EN.
With $\Delta=4$s (left), increasing the rewind length $k$ commits a shorter prefix and leaves a larger editable buffer, typically improving COMETKiwi at the cost of higher AL.
With $k=5$ (right), increasing chunk size $\Delta$ exposes more context per step and improves quality while increasing latency.
Together, $k$ and $\Delta$ provide practical knobs to tune the quality--latency frontier for no-revision streaming translation.

\begin{table}[t]
\resizebox{\linewidth}{!}{%

\centering
\small
\begin{tabular}{lcccc}
\toprule
System & IT$\rightarrow$EN & JA$\rightarrow$EN & PT$\rightarrow$EN & Avg. \\
\midrule
Phi-4-mini-MM        & 41.42 & 30.54 & 55.28 & 42.41 \\
Whisper large-v2     & 30.90 & 26.10 & 51.60 & 36.20 \\
GPT-4o               & 38.67 & 31.87 & 52.10 & 40.88 \\
\midrule
CSSEL offline        & 41.92 & 30.30 & 54.64 & 42.29 \\
CSSEL streaming      & 39.68 & 27.52 & 51.65 & 39.62 \\
CSSEL-P2P streaming  & 40.89 & 29.16 & 53.13 & 41.06 \\
\bottomrule
\end{tabular}
}
\caption{SacreBLEU results on a diagnostic CoVoST2 X$\rightarrow$EN subset. Streaming systems use cumulative chunk decoding with $\Delta=4$s and rewind length $k=5$.}
\label{tab:covost2_sacrebleu}
\end{table}

\subsection{Public-Set Diagnostic Evaluation}

Beyond the in-house conversational evaluation, we further conduct
a small diagnostic evaluation on CoVoST2\cite{wang2021covost} X$\rightarrow$EN using SacreBLEU\cite{post-2018-call}, to
check whether the same streaming behavior appears on a public
test set. We focus on the relative trend under an identical model
family and decoding setup.

As shown in Table~\ref{tab:covost2_sacrebleu}, moving from
offline CSSEL decoding to streaming decoding with $\Delta=4$s
and $k=5$ reduces the average BELU score from 42.29 to
39.62, reflecting the expected cost of prefix-only context and
committed-prefix decoding. P2P fine-tuning improves the
streaming average to 41.06, recovering more than half of the
offline-to-streaming degradation. The gains are consistent across
all three evaluated directions: +1.21 on IT$\rightarrow$EN,
+1.64 on JA$\rightarrow$EN, and +1.48 on PT$\rightarrow$EN.

\subsection{Discussion}
The results suggest that the main benefit of P2P supervision is behavioral:
it teaches the model to avoid premature commitments and to produce prefixes
that remain extendable as more speech arrives. This is complementary to
architectural approaches for SimulST. Rather than predicting explicit read/write
actions, our method encodes emission timing in the target prefixes themselves,
allowing the same decoder-only generation interface to be used at inference
time. The remaining limitations are the dependence on teacher-label quality and
the need to tune $\Delta$, $k$, and the waiting budget $K$ for different latency
requirements.

\section{Related Work}

Early work in simultaneous machine translation introduced the prefix-to-prefix formulation and controllable policies such as wait-$k$ \cite{ma2019stacl}, while adaptive approaches further learned read/write behavior through monotonic attention or related mechanisms \cite{arivazhagan2019monotonic,ma2019monotonic}. These ideas were later adapted from text to end-to-end simultaneous speech translation, where speech-specific pre-decision, segmentation, and latency evaluation introduce additional challenges \cite{ma2020simulmt,ma2020simuleval}. A related line of work explores whether offline sequence-to-sequence models can be used effectively in online settings. Liu et al. proposed partial hypothesis selection for chunk-based incremental speech recognition and translation, reducing latency by selecting stable partial outputs from growing speech chunks \cite{liu2020partial}. Pol\'ak et al. further showed that an offline speech translation model can be onlinized for simultaneous speech translation without modifying the original model \cite{polak2022cuni}. Re-translation methods also repeatedly translate growing input prefixes and allow revisions, often providing strong baselines when output instability is acceptable \cite{arivazhagan2020re,arivazhagan-etal-2020-translation}. In contrast, our work targets a stricter no-revision user interface: once a prefix is emitted, it is committed, while only a bounded suffix is internally rewound before commitment.

Another important direction is to improve the supervision signal for simultaneous translation. Since offline references are optimized for full-sentence translation, they can be poorly matched to low-latency generation, especially for language pairs with strong reordering. Prior work has addressed this issue by incorporating pseudo-references with fewer reorderings \cite{chen2021improving}. In  \cite{kano2022prefix}, the authors extracted alignments between bilingual prefix pairs and used them to segment streaming input and fine-tune a simultaneous translation model. Our method follows the same general motivation that simultaneous systems need prefix-level supervision, but differs in two aspects: the supervision is constructed with a teacher LLM under bounded waiting, and the student is trained to map partial speech directly to commit-ready translation prefixes, without using chunk transcriptions as inputs at inference time. Recent LLM-based SimulST systems have also introduced architectural or policy-aware designs, including blockwise-causal speech encoding \cite{ouyang2024fasst}, multi-turn unbounded speech translation with cache management \cite{ouyang2025infinisst}, and fully unidirectional architectures with explicit read/write prediction \cite{fu2025easist}. Our work is complementary to this trend: rather than adding policy heads, special read/write tokens, or new architectural pathways, we investigate whether fixed-chunk streaming decoding combined with teacher-labeled prefix-to-prefix supervision is sufficient to induce effective simultaneous speech translation behavior in a decoder-only speech LLM.

\section{Conclusion}
\label{sec:conclusion}

We propose a simple data-driven approach that combines fixed-chunk cumulative streaming decoding with rewind-based committed prefixes and teacher-labeled prefix-to-prefix supervision.
By fine-tuning CSSEL on commit-ready incremental targets, CSSEL-P2P achieves substantially better streaming quality at comparable latency on in-house conversational speech, with larger gains on reordering-heavy directions.
These results suggest that, under a robust fixed-chunk interface, prefix-to-prefix data construction and supervision can be a primary lever for enabling effective SimulST with decoder-only LLMs.
In future work, we will explore broader teacher strategies, more principled target construction, and scaling prefix-to-prefix supervision to more languages and domains.

\section{Generative AI Use Disclosure}
We used AI tools only for language polishing (grammar correction, spelling checks) and assisting in formatting and drawing LaTeX tables. All technical content, experiments, and conclusions were produced and verified by the authors.

\bibliographystyle{IEEEtran}
\bibliography{mybib}

@article{rousso2024tradition,
  title={Tradition or innovation: A comparison of modern ASR methods for forced alignment},
  author={Rousso, Rotem and Cohen, Eyal and Keshet, Joseph and Chodroff, Eleanor},
  journal={arXiv preprint arXiv:2406.19363},
  year={2024}
}

@article{xue2022large,
  title={Large-scale streaming end-to-end speech translation with neural transducers},
  author={Xue, Jian and Wang, Peidong and Li, Jinyu and Post, Matt and Gaur, Yashesh},
  journal={arXiv preprint arXiv:2204.05352},
  year={2022}
}

@inproceedings{rei2022cometkiwi,
  title={CometKiwi: IST-unbabel 2022 submission for the quality estimation shared task},
  author={Rei, Ricardo and Treviso, Marcos and Guerreiro, Nuno M and Zerva, Chrysoula and Farinha, Ana C and Maroti, Christine and De Souza, Jos{\'e} GC and Glushkova, Taisiya and Alves, Duarte and Coheur, Luisa and others},
  booktitle={Proceedings of the Seventh Conference on Machine Translation (WMT)},
  pages={634--645},
  year={2022}
}

@article{fu2025efficient,
  title={Efficient and adaptive simultaneous speech translation with fully unidirectional architecture},
  author={Fu, Biao and Yu, Donglei and Liao, Minpeng and Li, Chengxi and Chen, Yidong and Fan, Kai and Shi, Xiaodong},
  journal={arXiv preprint arXiv:2504.11809},
  year={2025}
}

@inproceedings{arivazhagan2019monotonic,
  title={Monotonic infinite lookback attention for simultaneous machine translation},
  author={Arivazhagan, Naveen and Cherry, Colin and Macherey, Wolfgang and Chiu, Chung-Cheng and Yavuz, Semih and Pang, Ruoming and Li, Wei and Raffel, Colin},
  booktitle={Proceedings of the 57th Annual Meeting of the Association for Computational Linguistics},
  pages={1313--1323},
  year={2019}
}

@inproceedings{ma2019stacl,
  title={STACL: Simultaneous translation with implicit anticipation and controllable latency using prefix-to-prefix framework},
  author={Ma, Mingbo and Huang, Liang and Xiong, Hao and Zheng, Renjie and Liu, Kaibo and Zheng, Baigong and Zhang, Chuanqiang and He, Zhongjun and Liu, Hairong and Li, Xing and others},
  booktitle={Proceedings of the 57th Annual Meeting of the Association for Computational Linguistics},
  pages={3025--3036},
  year={2019}
}

@inproceedings{arivazhagan2020re,
  title={Re-translation versus streaming for simultaneous translation},
  author={Arivazhagan, Naveen and Cherry, Colin and Macherey, Wolfgang and Foster, George},
  booktitle={Proceedings of the 17th International Conference on Spoken Language Translation},
  pages={220--227},
  year={2020}
}

@inproceedings{chen2021improving,
  title={Improving Simultaneous Translation by Incorporating Pseudo-References with Fewer Reorderings},
  author={Chen, Junkun and Zheng, Renjie and Kita, Atsuhito and Ma, Mingbo and Huang, Liang},
  booktitle={Proceedings of the 2021 Conference on Empirical Methods in Natural Language Processing},
  pages={5857--5864},
  year={2021}
}

@inproceedings{chen2021direct,
  title={Direct Simultaneous Speech-to-Text Translation Assisted by Synchronized Streaming ASR},
  author={Chen, Junkun and Ma, Mingbo and Zheng, Renjie and Huang, Liang},
  booktitle={Findings of the Association for Computational Linguistics: ACL-IJCNLP 2021},
  pages={4618--4624},
  year={2021}
}

@article{wang2021covost,
  title={CoVoST 2 and Massively Multilingual Speech Translation},
  author={Wang, Changhan and Wu, Anne and Gu, Jiatao and Pino, Juan},
  journal={Proc. Interspeech 2021},
  pages={2247--2251},
  year={2021}
}

@inproceedings{post-2018-call,
  title = "A Call for Clarity in Reporting {BLEU} Scores",
  author = "Post, Matt",
  booktitle = "Proceedings of the Third Conference on Machine Translation: Research Papers",
  month = oct,
  year = "2018",
  address = "Belgium, Brussels",
  publisher = "Association for Computational Linguistics",
  url = "https://www.aclweb.org/anthology/W18-6319",
  pages = "186--191",
}

@inproceedings{ren2020simulspeech,
  title={SimulSpeech: End-to-end simultaneous speech to text translation},
  author={Ren, Yi and Liu, Jinglin and Tan, Xu and Zhang, Chen and Qin, Tao and Zhao, Zhou and Liu, Tie-Yan},
  booktitle={Proceedings of the 58th Annual Meeting of the Association for Computational Linguistics},
  pages={3787--3796},
  year={2020}
}

@inproceedings{ma2020simuleval,
  title={SIMULEVAL: An evaluation toolkit for simultaneous translation},
  author={Ma, Xutai and Dousti, Mohammad Javad and Wang, Changhan and Gu, Jiatao and Pino, Juan},
  booktitle={Proceedings of the 2020 Conference on Empirical Methods in Natural Language Processing: System Demonstrations},
  pages={144--150},
  year={2020}
}

@article{achiam2023gpt,
  title={Gpt-4 technical report},
  author={Achiam, Josh and Adler, Steven and Agarwal, Sandhini and Ahmad, Lama and Akkaya, Ilge and Aleman, Florencia Leoni and Almeida, Diogo and Altenschmidt, Janko and Altman, Sam and Anadkat, Shyamal and others},
  journal={arXiv preprint arXiv:2303.08774},
  year={2023}
}

@article{touvron2023llama,
  title={Llama 2: Open foundation and fine-tuned chat models},
  author={Touvron, Hugo and Martin, Louis and Stone, Kevin and Albert, Peter and Almahairi, Amjad and Babaei, Yasmine and Bashlykov, Nikolay and Batra, Soumya and Bhargava, Prajjwal and Bhosale, Shruti and others},
  journal={arXiv preprint arXiv:2307.09288},
  year={2023}
}

@article{liu2024deepseek,
  title={Deepseek-v3 technical report},
  author={Liu, Aixin and Feng, Bei and Xue, Bing and Wang, Bingxuan and Wu, Bochao and Lu, Chengda and Zhao, Chenggang and Deng, Chengqi and Zhang, Chenyu and Ruan, Chong and others},
  journal={arXiv preprint arXiv:2412.19437},
  year={2024}
}

@article{team2025gemma,
  title={Gemma 3 technical report},
  author={Team, Gemma and Kamath, Aishwarya and Ferret, Johan and Pathak, Shreya and Vieillard, Nino and Merhej, Ramona and Perrin, Sarah and Matejovicova, Tatiana and Ram{\'e}, Alexandre and Rivi{\`e}re, Morgane and others},
  journal={arXiv preprint arXiv:2503.19786},
  year={2025}
}

@article{yang2025qwen3,
  title={Qwen3 technical report},
  author={Yang, An and Li, Anfeng and Yang, Baosong and Zhang, Beichen and Hui, Binyuan and Zheng, Bo and Yu, Bowen and Gao, Chang and Huang, Chengen and Lv, Chenxu and others},
  journal={arXiv preprint arXiv:2505.09388},
  year={2025}
}

@article{abouelenin2025phi,
  title={Phi-4-mini technical report: Compact yet powerful multimodal language models via mixture-of-{LoRA}s},
  author={Abouelenin, Abdelrahman and Ashfaq, Atabak and Atkinson, Adam and Awadalla, Hany and Bach, Nguyen and Bao, Jianmin and Benhaim, Alon and Cai, Martin and Chaudhary, Vishrav and Chen, Congcong and others},
  journal={arXiv preprint arXiv:2503.01743},
  year={2025}
}

@inproceedings{gulati20_interspeech,
  author={Anmol Gulati and James Qin and Chung-Cheng Chiu and Niki Parmar and Yu Zhang and Jiahui Yu and Wei Han and Shibo Wang and Zhengdong Zhang and Yonghui Wu and Ruoming Pang},
  title={Conformer: Convolution-augmented {Transformer} for Speech Recognition},
  year=2020,
  booktitle={Proc. Interspeech},
doi={10.21437/Interspeech.2020-3015},
}

@article{ouyang2024fasst,
  title={Fasst: Fast llm-based simultaneous speech translation},
  author={Ouyang, Siqi and Xu, Xi and Dandekar, Chinmay and Li, Lei},
  journal={arXiv preprint arXiv:2408.09430},
  year={2024}
}

@inproceedings{ouyang2025infinisst,
  title={Infinisst: Simultaneous translation of unbounded speech with large language model},
  author={Ouyang, Siqi and Xu, Xi and Li, Lei},
  booktitle={Findings of the Association for Computational Linguistics: ACL 2025},
  pages={3032--3046},
  year={2025}
}

@inproceedings{ma2020simulmt,
  title={SimulMT to SimulST: Adapting simultaneous text translation to end-to-end simultaneous speech translation},
  author={Ma, Xutai and Pino, Juan and Koehn, Philipp},
  booktitle={Proceedings of the 1st Conference of the Asia-Pacific Chapter of the Association for Computational Linguistics and the 10th International Joint Conference on Natural Language Processing},
  pages={582--587},
  year={2020}
}

@inproceedings{ma2021streaming,
  title={Streaming simultaneous speech translation with augmented memory transformer},
  author={Ma, Xutai and Wang, Yongqiang and Dousti, Mohammad Javad and Koehn, Philipp and Pino, Juan},
  booktitle={ICASSP 2021-2021 IEEE International Conference on Acoustics, Speech and Signal Processing (ICASSP)},
  pages={7523--7527},
  year={2021},
  organization={IEEE}
}

@article{chiu2017monotonic,
  title={Monotonic chunkwise attention},
  author={Chiu, Chung-Cheng and Raffel, Colin},
  journal={arXiv preprint arXiv:1712.05382},
  year={2017}
}

@article{ma2019monotonic,
  title={Monotonic multihead attention},
  author={Ma, Xutai and Pino, Juan and Cross, James and Puzon, Liezl and Gu, Jiatao},
  journal={arXiv preprint arXiv:1909.12406},
  year={2019}
}

@inproceedings{arivazhagan2020respeech,
  title={Re-translation strategies for long form, simultaneous, spoken language translation},
  author={Arivazhagan, Naveen and Cherry, Colin and Te, Isabelle and Macherey, Wolfgang and Baljekar, Pallavi and Foster, George},
  booktitle={ICASSP 2020-2020 IEEE International Conference on Acoustics, Speech and Signal Processing (ICASSP)},
  pages={7919--7923},
  year={2020},
  organization={IEEE}
}

@inproceedings{omachi2023align,
  title={Align, write, re-order: Explainable end-to-end speech translation via operation sequence generation},
  author={Omachi, Motoi and Yan, Brian and Dalmia, Siddharth and Fujita, Yuya and Watanabe, Shinji},
  booktitle={ICASSP 2023-2023 IEEE International Conference on Acoustics, Speech and Signal Processing (ICASSP)},
  pages={1--5},
  year={2023},
  organization={IEEE}
}

@inproceedings{liu2020partial,
  title     = {Low-Latency Sequence-to-Sequence Speech Recognition and Translation by Partial Hypothesis Selection},
  author    = {Liu, Danni and Spanakis, Gerasimos and Niehues, Jan},
  booktitle = {Proceedings of Interspeech 2020},
  year      = {2020},
  pages     = {3620--3624},
  doi       = {10.21437/Interspeech.2020-2897}
}

@inproceedings{polak2022cuni,
  title     = {{CUNI}-{KIT} System for Simultaneous Speech Translation Task at {IWSLT} 2022},
  author    = {Pol{\'a}k, Peter and Pham, Ngoc-Quan and Nguyen, Tuan Nam and Liu, Danni and Mullov, Carlos and Niehues, Jan and Bojar, Ond{\v{r}}ej and Waibel, Alexander},
  booktitle = {Proceedings of the 19th International Conference on Spoken Language Translation (IWSLT 2022)},
  year      = {2022},
  pages     = {277--285},
  address   = {Dublin, Ireland},
  publisher = {Association for Computational Linguistics},
  doi       = {10.18653/v1/2022.iwslt-1.24}
}

@inproceedings{kano2022prefix,
  title     = {Simultaneous Neural Machine Translation with Prefix Alignment},
  author    = {Kano, Yasumasa and Sudoh, Katsuhito and Nakamura, Satoshi},
  booktitle = {Proceedings of the 19th International Conference on Spoken Language Translation (IWSLT 2022)},
  year      = {2022},
  pages     = {22--31},
  address   = {Dublin, Ireland},
  publisher = {Association for Computational Linguistics},
  doi       = {10.18653/v1/2022.iwslt-1.3}
}

@inproceedings{arivazhagan-etal-2020-translation,
    title = "Re-translation versus Streaming for Simultaneous Translation",
    author = "Arivazhagan, Naveen  and
      Cherry, Colin  and
      Macherey, Wolfgang  and
      Foster, George",
    editor = {Federico, Marcello  and
      Waibel, Alex  and
      Knight, Kevin  and
      Nakamura, Satoshi  and
      Ney, Hermann  and
      Niehues, Jan  and
      St{\"u}ker, Sebastian  and
      Wu, Dekai  and
      Mariani, Joseph  and
      Yvon, Francois},
    booktitle = "Proceedings of the 17th International Conference on Spoken Language Translation",
    month = jul,
    year = "2020",
    address = "Online",
    publisher = "Association for Computational Linguistics",
    url = "https://aclanthology.org/2020.iwslt-1.27/",
    doi = "10.18653/v1/2020.iwslt-1.27",
    pages = "220--227"
}

@article{fu2025easist,
  title={Efficient and Adaptive Simultaneous Speech Translation with Fully Unidirectional Architecture},
  author={Fu, Biao and Yu, Donglei and Liao, Minpeng and Li, Chengxi and Chen, Yidong and Fan, Kai and Shi, Xiaodong},
  journal={arXiv preprint arXiv:2504.11809},
  year={2025}
}

\end{document}